\documentclass{article} 
\usepackage[utf8]{inputenc} 
\usepackage[T1]{fontenc} 
\usepackage[letterpaper]{geometry}
\usepackage{latexsym}
\usepackage[linesnumbered,ruled]{algorithm2e}
\usepackage{multirow}
\usepackage{fstc,times}
\usepackage{url}
\usepackage{amssymb}
\usepackage{amsmath}
\usepackage{graphicx}
\usepackage{booktabs}
\usepackage{caption}
\usepackage{subcaption}
\makeatletter
\newcommand{\@BIBLABEL}{\@emptybiblabel}
\newcommand{\@emptybiblabel}[1]{}
\makeatother
\usepackage[hidelinks]{hyperref}

\title{Few-Shot Text Classification with Pre-Trained Word Embeddings and a Human in the Loop}

\author{Katherine Bailey \and Sunny Chopra \\
  Acquia \\
  \texttt{\{katherine.bailey,sunny.chopra\}@acquia.com}
}
\date{}

\begin{document}
\maketitle

\begin{abstract}
Most of the literature around text classification treats it as a supervised learning problem: given a corpus of labeled documents, train a classifier such that it can accurately predict the classes of unseen documents. In industry, however, it is not uncommon for a business to have entire corpora of documents where few or none have been classified, or where existing classifications have become meaningless. With web content, for example, poor taxonomy management can result in labels being applied indiscriminately, making filtering by these labels unhelpful. Our work aims to make it possible to classify an entire corpus of unlabeled documents using a human-in-the-loop approach, where the content owner manually classifies just one or two documents per category and the rest can be automatically classified. This "few-shot" learning approach requires rich representations of the documents such that those that have been manually labeled can be treated as prototypes, and automatic classification of the rest is a simple case of measuring the distance to prototypes. This approach uses pre-trained word embeddings, where documents are represented using a simple weighted average of constituent word embeddings. We have tested the accuracy of the approach on existing labeled datasets and provide the results here. We have also made code available for reproducing the results we got on the 20 Newsgroups dataset\footnote{\url{https://github.com/katbailey/few-shot-text-classification}}.
\end{abstract}

\section{Introduction}

\subsection*{Word Embeddings}

Word Embeddings are representations of words as low-dimensional vectors of real numbers that capture the semantic relationships between words. In Natural Language Processing (NLP), some method for converting words to numeric values is always necessary as computation only works with numbers, not raw text. \citep{mikolov2013distributed} introduced efficient techniques for learning distributed vector representations of words from huge corpora. This method is called word2vec, and since then alternative approaches have been put forward by others such as \citep{pennington2014glove}, known as GloVe, and \citep{bojanowski2016subword}, known as FastText. There is also doc2vec for learning representations of entire documents.

These approaches differ in the way the representations are generated.  Word2vec and FastText are "predictive" models, whereas GloVe is categorized as a "count-based" model.  Predictive models learn their vectors by predicting the target word given neighboring context words using a feed-forward neural network. The weights of this network are optimized by using stochastic gradient descent, and these weights are vector representations of the words in the vocabulary. In contrast, count-based models learn their vectors by finding a lower dimensional representation of each word by minimizing a reconstruction loss function on the co-occurrence count matrix given as an input.

\subsection*{Few-Shot Learning}
Few-shot learning is an approach to classification that works with only a few human labeled examples. It often goes hand-in-hand with transfer learning, a technique involving learning representations during one task that can then be applied to a different task, because it is the richness of the learned representations that makes it possible to learn from just a few examples. The use of pre-trained word embeddings is an example of transfer learning. One-Shot learning is a special case of Few-Shot Learning: as the name suggests, it means learning to classify objects when only one labeled example exists per class.

\subsection*{Human-in-the-Loop}
The term Human-in-the-Loop (HitL) refers to any Machine Learning technique that involves human input in the training process. It includes techniques such as active learning, where humans handle low confidence predictions, as well as crowd-sourced approaches to labeling data sets. In our case, the humans perform the "few shots" in our few-shot learning system.

\section{Few-Shot Text Classification with a Human in the Loop}
Our approach involves a "classification engine" that the user (the content owner) interacts with. A batch of documents is fed to the engine and each document is converted into a 300-dimensional vector. A set of two or more categories is specified, and Latent Dirichlet Allocation is run on the batch using the number of categories provided as the number of topics. This serves to surface the most likely representative documents for each category. These are then presented to the user, who must choose some documents to manually classify for each category. Once this step is complete, the system has a vector representing each category: if only one document was classified for a category then that document's vector is used as the category vector, otherwise the vectors of multiple documents are averaged together. Once this is done, the remaining documents are compared against each category representative using simple cosine similarity and each one is assigned the category whose vector it is closest to. A score is also assigned for each prediction. All of these steps are explained in section \ref{approach}.

\section{Related work}
In \citep{arora2017asimple} the authors present an approach to representing sentences or entire documents as vectors. They first take the weighted average of the constituent (pre-trained) word vectors of the document. The weighting method serves to down-weight frequent words. They then run principal components analysis (PCA) on the batch, and the final embedding for each document is obtained by subtracting the projection of the set of sentence embeddings to their first principal component. The intuition behind this is that common methods, such as GloVe, for computing word vectors based on co-occurrence statistics lead to large components that contain no semantic information. A similar insight is presented in \citep{mu2017allbuttop}.

In \citep{snell2017prototypical}, the authors present the idea of \textit{Prototypical Networks} as a way of performing few-shot classification. In this approach, each class prototype is the mean vector of the vectorized support points belonging to its class and they are learned through gradient-descent-based training episodes on subsets of training examples. Classification then involves finding the nearest class prototype for each query vector. This is similar to how we perform classification but in our case we are actively seeking the best class prototypes by using a human-in-the-loop to choose them.

\section{Approach in Detail} \label{approach}
In this section, we provide the details of the text classification system.

\subsection{Document Embeddings}
 For any given collection of documents, which we refer to as a batch, each document needs to be converted to a fixed-length vector so that we can measure similarity between them. 

For our task, which is about classifying content, we found there to be little improvement in accuracy when running the PCA step, as suggested by \citep{arora2017asimple} and \citep{mu2017allbuttop}, over just using the weighted average of the word vectors. Morevover, PCA introduced a level of complexity that meant the embedding of a document was always batch-specific. For pragmatic reasons it was preferable for us to have document representations that were independent of the batch they came from. Hence our embedding method is simply:

\begin{algorithm}[!h]
\SetKwInOut{Input}{Input}
\SetKwInOut{Output}{Output}
\Input{Word representations $\{v_w : w\in \mathcal{V}\}$, a set of documents $ \mathcal{D}$, parameter $\alpha$, and estimated probabilities  $\{p(w) : w\in \mathcal{V}\}$ of the words}
\Output{Document embeddings $\{v_d : d\in \mathcal{D}\}$.}
\For {all documents d in $\mathcal{D}$}{
	$ v_d \leftarrow \frac{1}{|d|} \sum_{w\in d} \frac{\alpha}{\alpha + p(w)}v_w$
	}

\caption{Weighted average algorithm for document representations.}
\label{algo:representation}
\end{algorithm}

\subsection{Classification}
The classification task involves manually labelling a small number of documents in each class and using these as representatives of the class. In the one-shot case, this means that our representation of a class is simply the single document vector that has been labeled for that class. If we have more than one representative document per class, we simply take the average of those vectors as our category representative. Once we have a representative for each class, we can proceed to predict classes for the rest.

\begin{algorithm}[!h]
\SetKwInOut{Input}{Input}
\SetKwInOut{Output}{Output}
\Input{Class representative vectors $\{v_c : c\in \mathcal{C}\}$, a set of document vectors $\{v_d : d\in \mathcal{D}\}$}
\Output{Predicted class for each document.}
\For {all documents d in $\mathcal{D}$}{
	\For {all classes c in $\mathcal{C}$} {
	$ sims_c \leftarrow cosine\_similarity(v_d, v_c)$
	}
	$\hat{c_d} \leftarrow argmax(sims)$
}
\caption{Predicting classes using cosine similarity with class representative vectors.}
\label{algo:classification}
\end{algorithm}

\subsection{Surfacing good representatives}
Choosing good class representatives is of vital importance to the accuracy of this approach. The human-in-the-loop will be making the final choice about which documents to use as representatives for each class, but we need to make it as easy as possible to choose the best representatives. Our approach is to run Latent Dirichlet Allocation (LDA) on the batch of documents. This is a probabilistic approach to topic modeling which, given a number of topics, will infer those topics as distributions over the words in the vocabulary. Each document is assumed to be some mixture of topics, and these are assigned probabilities based on the words contained in the document. For example document 1 could be deemed to be 90\% about topic A and 10\% about topic B. If you assume that the topics are mutually exclusive then it is reasonable to simply assign each document to the topic it has the highest probability of belonging to. We perform this step, and then for each inferred topic we rank the documents assigned to it in descending order of their probability of belonging to that topic. The assumption here is that LDA, in inferring topics, has approximately teased apart the actual categories, and so the idea is to have an ordering of the documents within a batch such that the first page of documents seen in the UI is likely to have a mix of good representatives for each category.

\section{Experiments}
We evaluate the accuracy of performing one-shot classification on some publicly available labeled datasets.

\subsection{Datasets}
We created 2- and 3-category subsets of the "20 Newsgroups" dataset. The original dataset consists of 20,000 messages taken from 20 newsgroups on subjects like cars, religion, guns and baseball. We used the Scikit-learn python library to extract subsets of this data. We also created 2-, 3- and 4-category subsets of the DBPedia dataset, which contains the first paragraph of the Wikipedia page for around half a million entities in 15 categories, e.g. Animal, Film, Plant, Company, etc.

We perform some basic cleaning of the text and remove stop words before converting documents into vectors.

\begin{table}[]
\centering
\captionsetup{position=bottom}
\begin{tabular}{lll}
\toprule
Categories                  & \# docs & Max acc. \\
\midrule
Village, Film                & 9307         & 0.9992       \\
Village,Animal              & 8960         & 0.9979       \\
Animal, Film                 & 8947         & 0.9946       \\
Animal, Company              & 9220         & 0.9943       \\
Animal, Film, Company, Village & 18527        & 0.9706       \\
autos, baseball              & 1046         & 0.9703       \\
guns, hardware               & 1078         & 0.9693       \\
mideast, electronics         & 1063         & 0.9689       \\
Animal, Film, Company         & 13867        & 0.9688       \\
christian, guns              & 1099         & 0.9380       \\
med, electronics             & 1112         & 0.9324       \\
atheism, space               & 1004         & 0.9232       \\
baseball, hockey             & 1069         & 0.9063       \\
autos, baseball, space        & 1605         & 0.8964       \\
Animal, Plant                & 8730         & 0.8540       \\
politics, religion           & 774          & 0.8264      \\
\bottomrule
\end{tabular}
\caption{Maximum achieved accuracy using one-shot classification}\label{bruteforce}
\end{table}

\subsection{Maximum One-Shot Accuracy}
In order to know how accurate our classification approach can be in theory (i.e. if the user happens to be lucky enough to find the best representatives for each category) we did some brute force testing of the approach, where we went through thousands of combinations of document representatives to see what the highest achievable accuracy was with one-shot learning. In some cases, i.e. where there were only two categories and only around 500 documents per category, we did an exhaustive search, testing all possible combinations. In other cases we randomly sampled from all possible combinations. Table \ref{bruteforce} shows the maximum accuracy we achieved on various subsets of categories.

As expected the accuracy is high when the categories are very distinct, such as the "Animal" and "Film" categories in the DBPedia dataset: there would be very little overlap in the types of words used in articles about animals vs articles about films. We looked at the few misclassifications that arose when the best known representatives were used in this dataset and found that the films misclassified as animals were films about animals and the animals misclassified as films were all thoroughbred racehorses that had won or been nominated for awards.

These results show us that for a batch of documents to be classified into separate categories, it is possible to achieve very high accuracy with just a single labeled example per category, provided that 1) those categories truly are reflected in the words of the documents and 2) good enough representatives are chosen for each category. Since for any given batch we have no control over how well the words in the documents reflect the desired categories, the problem we focus on is how to choose the best representatives.

\subsection{Results with LDA}
We ran LDA on our datasets, specifying the number of known categories as the number of topics, and then tested the maximum accuracy that could be obtained using one-shot learning when trying only the possible combinations of representatives in the top 12 documents surfaced by LDA (as described in section \ref{approach}). The idea is that the user will be presented with documents for classification, perhaps 12 documents per page, and ideally they wouldn't have to look beyond the first, or at most the second, page of documents in order to choose representatives for each category.

\begin{table}[]
\centering
\captionsetup{position=bottom}
\begin{tabular}{lll}
\toprule
Categories                  & Max acc. & LDA max \\
\midrule
Village, Film                & 0.9992       & 0.998388    \\
Village, Animal              & 0.9979       & --         \\
Animal, Film                 & 0.9946       & 0.992510    \\
Animal, Company              & 0.9943       & 0.987633    \\
Animal, Film, Company, Village & 0.9706       & --         \\
autos, baseball              & 0.9703       & 0.945402    \\
guns, hardware               & 0.9693       & 0.965613    \\
mideast, electronics         & 0.9689       & 0.959472    \\
Animal, Film, Company         & 0.9688       & 0.961771    \\
christian, guns              & 0.9380       & 0.925251    \\
med, electronics             & 0.9324       & 0.931532    \\
atheism, space               & 0.9232       & 0.864271    \\
baseball, hockey             & 0.9063       & 0.798500    \\
autos, baseball, space        & 0.8964       & 0.803371    \\
Animal, Plant                & 0.8540       & --         \\
politics, religion           & 0.8264       & 0.804404   \\
\bottomrule
\end{tabular}
\caption{Maximum brute-force accuracy obtained and maximum accuracy obtained using only combinations of the top 12 LDA-surfaced documents}\label{ldaacc}
\end{table}

In table \ref{ldaacc} we show the same maximum accuracy obtained via the brute-force testing of combinations along with the maximum accuracy achieved by testing the combinations from the top 12 documents. In many cases the LDA accuracy is quite close to the maximum brute-force accuracy, however in 3 cases we could not get a result for LDA because the top 12 documents did not include representatives from all categories. This is perhaps understandable in the case of the 4-category dataset, "Animal, Film, Company, Village." In the case of the "Animal, Plant" dataset we see that the brute force accuracy is relatively low, due to the fact that many similar words would be used to describe animals and plants ("species", "family", "habitat", etc), and so this perhaps explains why LDA had a difficult time teasing apart the topics. However in the case of the "Village,Animal" dataset, which gets a maximum brute-force accuracy of 99.79\% the same argument clearly cannot be made. This tells us that there is much room for improvement in our method for inferring topics in datasets for the purpose of surfacing good category representatives.

\begin{table}[]
\centering
\captionsetup{position=bottom}
\begin{tabular}{lll}
\toprule
Categories                      & GloVe LDA       & FastText LDA \\
\midrule
Village, Film                               & 0.9984       & 0.9932                 \\
Animal, Film                                & 0.9925       & 0.9934                 \\
MeansOfTransportation, NaturalPlace          & 0.9922       & 0.9926                 \\
Village, Company                            & 0.9895       & 0.9896                 \\
Animal, Company                             & 0.9876       & 0.9905                 \\
Building, NaturalPlace                      & 0.9791       & 0.9807                 \\
Album, Film                                 & 0.974        & 0.9825                 \\
guns, hardware                              & 0.9656       & 0.9638                 \\
Film, Company                               & 0.9632       & 0.9664                 \\
Animal, Film, Company                        & 0.9618       & 0.9445                 \\
Artist, Athlete                             & 0.9604       & 0.922                  \\
mideast, electronics                        & 0.9595       & 0.9746                 \\
autos, baseball                             & 0.9454       & 0.9511                 \\
autos, hockey                               & 0.9413       & 0.9376                 \\
Building, EducationalInstitution            & 0.9345       & 0.935                  \\
Film, WrittenWork                           & 0.9341       & 0.9376                 \\
med, electronics                            & 0.9315       & 0.9153                 \\
christian, guns                             & 0.9253       & 0.9243                 \\
med, religion                               & 0.888        & 0.888                  \\
atheism, space                              & 0.8643       & 0.8713                 \\
autos, guns                                 & 0.8417       & 0.8407                 \\
autos, electronics                          & 0.8291       & 0.845                  \\
space, med                                  & 0.8111       & 0.8405                 \\
politics, religion                          & 0.8044       & 0.7526                 \\
autos, baseball, space                       & 0.8034       & 0.8414                 \\
baseball, hockey                            & 0.7985       & 0.8594                 \\
religion, mideast                           & 0.7526       & 0.7808                 \\
christian, atheism                          & 0.6553       & 0.6777                 \\
atheism, religion                           & 0.6304       & 0.6522                 \\
\bottomrule
\end{tabular}
\caption{Max accuracies on combinations of top 12 LDA documents: GloVe vs FastText}\label{lda_fasttext}
\end{table}

We also tested the LDA accuracy on some datasets for which we had not run a brute-force test. Table \ref{lda_fasttext} shows these results, tested using both GloVe and FastText pre-trained word embeddings.

\subsection{Relative length of category representatives}
We were curious about the extent to which relative length of the chosen category representatives mattered. For example, if the user chose a very short document as the sole representative for category A and a very long document, or multiple documents, for category B, how much would this skew the predictions towards category B. On one dataset only, the 2-category subset "autos, baseball" from 20 Newsgroups, we went through every possible combination of category representatives, noted the length of each document, ran the auto-classification step, and noted the number of predictions in each category. With this information, we were able to examine the correlation between relative representative length and relative prediction count for that category. We found a strong positive correlation between them, roughly 0.8. This means we need to either provide guidance to the user when selecting representatives (to make sure they select representatives of roughly equal length), or we need to filter the documents we present to the user to remove very long and very short ones.

\section{Discussion}
Our testing on the 20 Newsgroups and DBPedia datasets showed that the general approach is sound, that the document embeddings using weighted averages of pre-trained word embeddings are useful representations, and that if good category representatives are chosen, high levels of accuracy can be achieved on the classification task. It also showed that LDA can help in choosing good representatives. The datasets we tested on were quite artificial: mostly just 2-category datasets chosen from larger datasets with more categories. However they did help identify certain characteristics that may make a real dataset suitable to use this approach on. The most important characteristic is that the categories must be sufficiently distinct and the words used in the documents should reflect those categories. The length of the documents chosen as representatives is also important - they should be roughly equal in length.

As to the question of working with more categories, why not test on the entire 20 Newsgroups or DBPedia dataset? Brute-force testing becomes infeasible with larger numbers of categories, and so it wasn't possible to get a maximum achievable accuracy for the entire dataset (the maximum achieved accuracy of 97\% on a particular 4-category subset of the DBPedia dataset was based on testing 390,625 of the 625 trillion possible combinations). In any case, it is unclear whether this approach will be appropriate for datasets with many categories --- say, more than four or five --- seeing as it requires the human in the loop to find good representatives for each one, a task which might prove daunting in such a case even if the category breakdown has been successfully approximated through LDA or similar. We feel that our text classification approach is best suited to 2- or 3-category classification tasks. An interesting use case for 2-category, i.e. binary, classification might be topic stance detection. Given user-generated content on a particular topic, e.g. posts on a web forum, it might be the case that different words are used depending on where the writer stands on that topic. An example of this would be some people using the term "family reunification" and others using "chain migration" to refer to the same thing. In this case our approach could be used to detect the stance reflected in each post.

\section{Future work}
The crucial step in our method is to present the user with suitable candidates to be labeled as representatives. We used Latent Dirichlet Allocation (LDA) for topic inference and found that while in many cases we got close to the maximum accuracy achieved through brute-force trials, in some cases it fell far short or even failed to tease apart the topics at all.

An alternative to LDA would be to use probabilistic Latent Semantic Analysis (pLSA) which treats topics as word distributions and uses probabilistic methods similar to LDA. But the use of Dirichlet priors in LDA for the document-topic and topic-word distributions in order to prevent over-fitting seems to make it a better choice. Our goal in the future is to improve the topic inference step, and so we will look to other alternatives. One idea is to use guided LDA as suggested in \citep{conf/eacl/JagarlamudiDU12}.  The seeds for guiding it will be the category names that the user specifies to classify the batch of documents.  Another approach is to use Gaussian LDA as proposed in \citep{conf/acl/DasZD15}. This approach is a good fit as it works with vectorized representations of words and documents. Yet another option would be to use ProdLDA as described in \citep{2017arXiv170301488S}, which is a neural network version in which the distribution over individual words is a product of experts rather than the mixture model used in LDA. Running clustering algorithms on the document representations is another approach we plan to try in order to solve the topic inference problem.

Although we only tested our approach on single labeled datasets, we would like to be able to apply it to multi-labeled datasets. One idea would be to use each label independently and do a binary classification of whether a document has that label or not. By treating each label independently of the other labels, we convert it into a single labeled problem. However, our current LDA approach to topic inference will not work in this case as it will always suggest the same ordering of documents regardless of which label the user is choosing representatives for. This is where guiding the LDA with the seeds of category names can prove immensely useful. An alternative approach, rather than treating each label independently, would be to use classifier chains for multi-label classification as suggested in \citep{Read2009}. This method may help us achieve a fair improvement in accuracy. The foremost issue again is the topic inference, which will be essential for accurate multi-label classification, and that's going to be our principal focus of research in the future.

\bibliography{fstc}
\bibliographystyle{iclr2017_conference}

\end{document}